\documentclass{article}

\PassOptionsToPackage{numbers, compress}{natbib}

\usepackage[preprint]{nips_2018}

\usepackage[utf8]{inputenc} 
\usepackage[T1]{fontenc}    
\usepackage{hyperref}       
\usepackage{url}            
\usepackage{booktabs}       
\usepackage{amsfonts}       
\usepackage{nicefrac}       
\usepackage{microtype}      

\usepackage{color}
\usepackage{amsmath}
\usepackage{relsize}
\usepackage{tabularx}
\usepackage{graphicx}
\usepackage{wrapfig}
\usepackage{multirow}
\usepackage{multicol}

\title{Wall Stress Estimation of Cerebral Aneurysm based on Zernike Convolutional Neural Networks}

\author{
  Zhiyu Sun \\
  University of Iowa, United States\\
  \texttt{zhiyu-sun@uiowa.edu} \\
  \And
    Jia Lu \\
    University of Iowa, United States \\
    \texttt{Jia-Lu@uiowa.edu} \\
    \AND
    Stephen Baek\\
    University of Iowa, United States \\
    \texttt{stephen-baek@uiowa.edu} \\
}

\begin{document}

\maketitle

\begin{abstract}
    Convolutional neural networks (ConvNets) have demonstrated an exceptional capacity to discern visual patterns from digital images and signals. Unfortunately, such powerful ConvNets do not generalize well to arbitrary-shaped manifolds, where data representation does not fit into a tensor-like grid. Hence, many fields of science and engineering, where data points possess some manifold structure, cannot enjoy the full benefits of the recent advances in ConvNets. The aneurysm wall stress estimation problem introduced in this paper is one of many such problems. The problem is well-known to be of a paramount clinical importance, but yet, traditional ConvNets cannot be applied due to the manifold structure of the data, neither does the state-of-the-art geometric ConvNets perform well. Motivated by this, we propose a new geometric ConvNet method named ZerNet, which builds upon our novel mathematical generalization of convolution and pooling operations on manifolds. Our study shows that the ZerNet outperforms the other state-of-the-art geometric ConvNets in terms of accuracy.
\end{abstract}

\section{Introduction}
Many areas of science and engineering have to deal with geometry data of some kind, because shape often encompasses a great deal of critical information for understanding phenomena. Such problems often boil down to a question of identifying latent geometric patterns behind a variety of shapes and correlating the patterns with a certain physical phenomena. In this regard, the convolutional neural networks, also known as CNN or \textit{ConvNets}, have demonstrated a phenomenal capacity in capturing and recognizing important visual features from images or signals, conceiving the unprecedented advances in machine learning and artificial intelligence research. 

However, like what has already been pointed out by Bronstein \textit{et al.}~\cite{bronstein2017geometric}, geometric features on free-form surfaces are not quite trivial to describe numerically and, hence, do not enjoy the luxury of such powerful ConvNets. That is, in contrast to plain signals or images, where a tensor-like grid structure is readily available, free-form surfaces, or more formally non-Euclidean manifolds, do not possess such grid structure, rendering significant difficulties in consistent data representation. In this reason, many of the core operations of ConvNets, including the very notion of function convolution, cannot be nicely generalized, which is the current bottleneck preventing the wide-spread use of ConvNets in computational geometry and many other relevant research areas.

The problem we are addressing in this paper, estimating the wall stress distribution on a cerebral aneurysm surface, is one of such problems that cannot be addressed using the traditional ConvNets. We hypothesize that the magnitude of the wall stress applied to cerebral aneurysms has a correlation with the local surface features of an aneurysm and the semantic context of how these features are aligned. If this hypothesis is true, the magnitude of the wall stress on an aneurysm could be estimated fairly quickly based on its geometric shape. As such, the use of computationally-heavy, finite element type of stress analysis methods will become unnecessary, which presents tons of clinical advantages in patient care. However, for the testing of this hypothesis, a statistical model that concerns with surface features needs to be built, which, unfortunately, is not possible because of the manifold structure of data as previously mentioned.

Therefore, we introduce in this paper a novel concept of Zernike convolution on manifold surfaces in an attempt to rigorously expand the concept of ConvNets to surface manifolds. Although we are targeting at a rather specific application of aneurysm wall stress estimation, we keep the generality in formulating the mathematical concepts and theorems behind our method, such that it is seamlessly scalable to other problems. The main scientific contributions of our work can be summarized as follows:
\begin{itemize}
  \item A novel concept of Zernike convolution generalizing the (planar) convolution operation is proposed. In particular, by using the Zernike polynomials as the function bases, the tensor field defined on the manifold is locally approximated. By proving that the function convolution is equivalent to the inner product between the Zernike coefficient, a simple and efficient convolution operation is rigorously defined on manifold domains.
  \item Compared with the current state-of-the-art methods for manifold convolution \cite{masci2015geodesic,boscaini2016learning, monti2017geometric}, our approach comes with a lesser number of neural weights as we demonstrated in Section \ref{Comparison}, presenting benefits to the computational efficiency.
  \item We propose ZerNet: a neural network architecture that generalize ConvNets to solve feature-based regression problems correlated to geometric properties residing in a manifold structure, showing a superior accuracy compared with the other state-of-the art geometric ConvNets.
\end{itemize}

\section{Related works}
\label{relatedworks}
One of the pioneering works in this area is due to Bruna \textit{et al.} \cite{bruna2013spectral,henaff2015deep}. They generalized the convolution operation \cite{lecun1998gradient} to graph-structured data by representing the graph convolution operation in terms of spectral bases of the graph-Laplacian. Defferrard \textit{et al.} \cite{defferrard2016convolutional} shares the similar idea of using the spectral bases for the definition of graph convolution. However, they introduced the Chebyshev polynomials as to avoid the explicit computation of the graph-Laplacian eigenbases which can be highly time consuming for large scale problems. Later, Kipf \textit{et al.}~\cite{kipf2016semi} further improved this strategy by defining more concise filters based only on the one-ring neighborhoods of the graph. In the meantime, there also have been a branch of research that attempts to generalize ConvNets based on spatial formulation \cite{duvenaud2015convolutional,niepert2016learning,hechtlinger2017generalization}, instead of using spectral bases. Those spatial approaches generalize the convolution as an inner product of the parameters (filter coefficients) with spatially close neighbors relying on the graph's spatial structure. Aforementioned graph-based generalization of ConvNets act as paradigms in attempts to generalize ConvNets to Non-Eulicidean structured data, providing some insights toward the application of ConvNets to surface geometry data.

More with the direct connection to surface geometry processing, several manifold-based approaches have also been reported in this area. Similar to the spectral formulations above, Boscaini \textit{et al}. \cite{boscaini2015learning} utilized the windowed Fourier transform \cite{shuman2016vertex} to define the convolution on localized spectral bases of a manifold. However, a fundamental issue of the spectral formulation type of convolution is its domain-dependent. The spectral filters learned with respect to the Fourier basis on one domain cannot be applied to another domain in a trivial way. Although, in the work \cite{boscaini2015learning}, some spatial interpretation on manifold is given by the localized spectral filters, which to some extent addressed such issue. The shift-invariance, a critical property should be preserved through the convolution operation, was still an unaddressed problem.

Therefore, an alternative approach generalizing the convolution purely in the spatial domain on manifolds ascends the stage. Masci \textit{et al}. \cite{masci2015geodesic} proposed Geodesic CNN (GCNN), the first version of spatial formulated geometric ConvNets on manifolds. In \cite{masci2015geodesic}, the filters were applied to the extracted patch which is represented as a combination of gaussian weights defined on a local geodesic polar coordinates. As a follow-up work, Boscaini \textit{st al}. \cite{boscaini2016learning} adopted an alternative approach of local patch construction by using anisotropic heat kernels \cite{andreux2014anisotropic}. Monti \textit{et al}. \cite{monti2017geometric} further generalized the patch construction by introducing some learnable parameters to parametrize the weighting function defined on a local system of pseudocoordinates. The strength of spatial based ConvNets on manifolds is the ability to adapt the model across different domains, in the mean time preserving the property of shift-invariance during the convolution operation. This property is crucial for the applications in geometry processing, satisfying the requirement that a ConvNet model trained on one shape can be applied to another one.

\section{Our approach}
In this section, we introduce a new concept of Zernike convolution to generalize ConvNets on manifolds. A conventional convolution operation defined on image domains amounts to extracting a local patch of pixels and taking the weighted sum over a template, or a convolutional filter. The inherent grid-like structure of image domains (i.e. Euclidean) presents a consistent global parametrization property, which allows the application of a template in a consistent orientation across the pixels. This, however, is not necessarily true on manifold domains where the Euclidean grid-structure is not available. Hence, the main challenge in this regard is how to formulate a strategy to apply convolutional filters in a consistent but effective manner.

\begin{figure}[h]
    \centering
    \includegraphics[width=\textwidth]{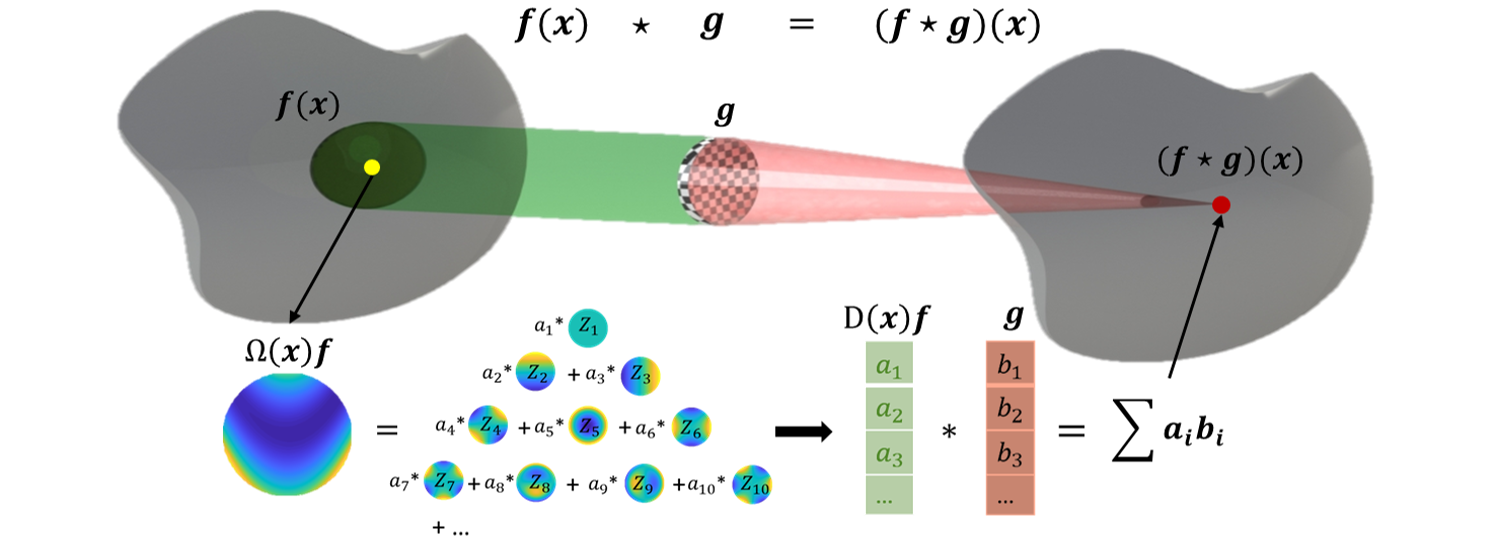}
    \caption{Zernike convolution. A local patch is approximated by the Zernike polynomial bases. Under this setting, function convolution becomes a simple dot product between the Zernike coefficients, which allows a simple but rigorous generalization of the convolution operation on manifolds.}
    \label{fig: Zernike Convolution diagram}
\end{figure}

To this end, we propose the following approach. First, for each point sampled on a given manifold, we define a local patch around the sampled point. Then, at each point, the tensor field defined on the manifold is locally approximated by using the Zernike polynomials as the function bases (Figure~\ref{fig: Zernike Convolution diagram}). Under this setting, we prove that the function convolution becomes merely the inner product between the Zernike coefficient, permitting a simple and efficient computation. Further, under this setting, the change of patch orientation can be achieved by simply multiplying a typical 2$\times$2 planar rotation matrix, which facilitates the angular operations (e.g. angular pooling) that mitigate the global parameterization issue.

\subsection{Zernike polynomials}
By definition, Zernike polynomials \cite{von1934beugungstheorie} are an orthogonal polynomial sequence (i.e. $<Z_i, Z_j> = \int{Z_i(t)Z_j(t)dt} = 0$ for $i \neq j$) defined over a unit disk. The Zernike polynomials are separated into even and odd polynomials denoted as $Z_n^{m}(r,\theta)$ and $Z_n^{-m}(r,\theta)$ respectively:
\begin{align}
    \begin{split}
        Z_n^{m}(r,\theta) &= R_n^{m}(r) \cos(m \theta), \\
        Z_n^{-m}(r,\theta) &= R_n^{m}(r) \sin(m \theta).
    \end{split}
\label{eq:Zernike}
\end{align}
where $m$ and $n$ are non-negative integer indices with $n\geq m$; $r\subseteq [0,1]$ is the radial distance; and $\theta$ is the azimuthal angle. Furthermore, the Zernike radial polynomial $R_n^{m}(r)$ is defined as:
\begin{equation}
    R_n^{m}(r) = 
    \begin{cases}
    \mathlarger{\sum}_{k=0}^{\frac{n-m}{2}}
    {\frac{(-1)^{k}(n-k)!}{k!(\frac{n+m}{2}-k))!(\frac{n-m}{2}-k)!}r^{n-2k}}, & \text{if } n-m \text{ even}\\
    0, & \text{otherwise}
    \end{cases}
    \label{eq:Zernike radial poly}
\end{equation}
In order to achieve a sequence of orthonormal bases on a unit disk, Zernike polynomials can be further normalized with a normalization factor $\sqrt{\frac{2-\delta(m,0)}{\pi}}$:
\begin{equation}
    {{\hat{Z}_n^{m}}(r,\theta)} = Z_n^{m}(r,\theta) \sqrt{\frac{2-\delta(m,0)}{\pi}},
    \label{eq:noramlized Zernike func}
\end{equation}
where $\delta$ denotes the Kronecker delta. 

The normalized Zernike polynomials are useful for decomposing complex functions or data due to their orthonormality over the unit disk. Any function $f(r,\theta)$ defined on the domain $[0,1]\times[0,2\pi)$ can be expressed as a sum of Zernike modes:
\begin{equation}
    f(r,\theta) = \sum_{n=0}^{\infty}\sum_{m=-n}^{n}a_{nm}  {{\hat{Z}_n^{m}}(r,\theta)}.
    \label{eq:zernike decomposition} 
\end{equation}
For simplicity, for the rest of the paper we use the term "\textit{Zernike functions}" standing for normalized Zernike polynomials, denoted as $Z_i(r, \theta)$ with index $i$ corresponding to a certain pair of $(n,m)$ in (\ref{eq:zernike decomposition}), then (\ref{eq:zernike decomposition}) can be expressed as:
\begin{equation}
    f(r,\theta) = \sum_{i=1}^{\infty}a_{i}{Z_i(r,\theta)}
    \label{eq:zernike approximation} 
\end{equation}

\subsection{Zernike convolution}
\label{Zernikeconvolution}
Building upon the above definition of (normalized) Zernike polynomials, we define key building blocks of Zernike convolutional neural networks. On a manifold domain $\mathcal{M}$, a convolutional response of a function $f$ at a point $x\in\mathcal{M}$ with respect to a convolutional filter $g$ is defined as follows:
\begin{equation}
    (f\star g)(x) = \int_{t\in\mathcal{M}} f(x) g(t, x) d\mathcal{M},
    \label{eq:manifold convolution}
\end{equation}
where $f$ and $g$ are functions defined on the manifold $\mathcal{M}$ and $g(\cdot, c)$ denotes a convolutional filter centered at $c$.

For a local extrinsic patch extraction on the manifold, we adopt a method similar to \cite{masci2015geodesic}. For a local neighborhood $\Omega(x)$ of $x\in\mathcal{M}$, a geodesic ball $B_{r_0}$ of radius $r_0$ is defined around $x$. Then, given that the radius $r_0$ is sufficiently small w.r.t. the local convexity radius of the manifold, the geodesic ball is homeomorphic to the topological disk, and thus, one can define a bijective mapping from $\Omega(x)$ to a unit disk. That is, there always exists a mapping $\chi: B_{r_0}(x) \mapsto [0,1] \times [0, 2\pi)$ in the polar coordinate convention. This permits the mapping of a function $f(x)$ on $\mathcal{M}$ to a function $(\Omega(x)f)(r,\theta)$ on the polar coordinates, as well as the convolutional kernel $g$:
\begin{equation}
    (\Omega(x)f)(r,\theta) = \sum_{i=1}^{\infty}a_{i}{Z_i(r,\theta)},
    \label{eq:f_Zernike}
\end{equation}
\begin{equation}
    (\Omega(x)g)(r,\theta) = \sum_{j=1}^{\infty}b_{j}{Z_j(r,\theta)},
    \label{eq:g_Zernike}
\end{equation}
with different sets of coefficients $\{a_i\}$ and $\{b_j\}$ respectively. Under this notion, the convolution operation on the manifold is defined as:
\begin{equation}
    (f\star g)(x) = \sum_{ij}^{\infty}{a_i b_j} \int_{\theta} \int_{r}{Z_i(r,\theta)}{Z_j(r,\theta)} r\ drd\theta.
    \label{eq:Zernik convolution 2}
\end{equation}
However, due to the orthonormality property of the \textit{Zernike functions} over the unit disk, (\ref{eq:Zernik convolution 2}) can further be simplified as:
\begin{equation}
    (f\star g)(x) = \sum_{i}^{\infty}{a_i b_i},
    \label{eq:Zernik convolution 3}
\end{equation}
which indicates that the convolution operation defined on the manifold is equivalent to the dot product of the Zernike coefficients $\{a_i\}$ and $\{b_i\}$.

\paragraph{Angular operation} 
In order to overcome the lack of an orientation-consistent global parametrization, we define the Zernike angular pooling operation in a similar spirit to \cite{masci2015geodesic}. From the definition of Zernike polynomials in (\ref{eq:f_Zernike}), A Zernike patch after undergoing a rotation $\triangle\theta$ can be represented as:
\begin{equation}
    (\Omega(x)f)(r,\theta+\triangle\theta) = \sum_{\text{even}}a_n^m\  Z_n^m(r,\theta+\triangle\theta) + \sum_{\text{odd}}a_n^{-m}\  Z_n^{-m}(r,\theta+\triangle\theta).
    \label{eq: Angular operation}
\end{equation}
However, it is easy to show from the trigonometric sum and product formulae that
\begin{equation}
    \left\{
    \begin{array}{l}
     Z_n^{-m}(r,\theta+\triangle\theta)\\
     Z_n^{m}(r,\theta+\triangle\theta)
    \end{array} 
    \right\}
    =
    \left\{
    \begin{array}{l}
     Z_n^{m}(r,\theta)cos(m \triangle\theta)-Z_n^{-m}(r,\theta)sin(m \triangle\theta)\\
     Z_n^{-m}(r,\theta)cos(m \triangle\theta)+Z_n^{m}(r,\theta)sin(m \triangle\theta)
    \end{array} 
    \right\}.
    \label{eq:Zernike angle funcs }
\end{equation}
Therefore, the function $\Omega(x)f$ after the rotation $\triangle\theta$ is merely a $2\times2$ rotational transform of the Zernike coefficients $a_n^m$ and $a_n^{-m}$ for all pairs of $(n, m)$:
\begin{equation}
    T_m = \begin{bmatrix}
        cos(m \triangle\theta) & -sin(m \triangle\theta) \\
        sin(m \triangle\theta) & cos(m \triangle\theta)
     \end{bmatrix}.
\end{equation}
Thus, in the Zernike convolution setting, the angular pooling operation can be achieved with any desired angle steps without losing any mathematical rigor, as opposed to the fixed angular bin heuristics as in \cite{masci2015geodesic}.
\paragraph{Patch operator}
We introduce a term called \textit{Zernike patch operator} $D(x)f$:
\begin{equation}
    D_i(x)f =a_i= \int_\theta\int_{r} (\Omega(x)f)(r,\theta) Z_i(r,\theta)r\ dr\ d\theta,
    \label{eq:Zernike patch operator}
\end{equation}
which is used to describe the operation that derives the set of coefficients $\{a_i\}$ in (\ref{eq:f_Zernike}). We can regard $D(x)f$ as an extracted intrinsic 'patch' on the manifold through aforementioned operation.

By introducing \textit{Zernike patch operator} and \textit{Zernike angular rotator}, we define the \textit{Zernike convolution}: 
\begin{equation}
    (f\star g)(x) = \sum_{i=1}^{N} (R(\triangle\theta)g_i)\ D_i(x)f,
    \label{eq:Zernike convolution 4}
\end{equation}
where $g$ denote the filters applied on the extracted 'patch' on the manifold, which can be rotated by arbitrary angle $\triangle\theta$ to resolve the issue of angular coordinate ambiguity, $N$ denote a finite number of Zernike functions used for decomposition.

\subsection{ZerNet}
With the convolution operation on manifold, \textit{Zernike convolution}, being well defined, now we are ready to extend ConvNet for geometric feature learning on 3D graphical models. We propose a framework of constructing Zernike convolutional neural network (ZerNet), which enables extracting geometric features from 3D graphical models and learning the correlations between such features and target properties distributed over the graphical models. ZerNet can take any geometry vectors distributed over the model surface (e.g. $XYZ-$ coordinates of mesh vertices) as input; finally outputs the target properties as a regressed scalar field. As the major building blocks of ZerNet, we introduce the following types of layers:

\paragraph{Zernike-patch extraction} layer is set as a fixed layer before \textit{Zernike convolution} layer, which typically takes the initial geometric vectors from input layer or the evolved feature vectors from previous layer as the input. In this layer, such input vectors are partitioned to the positions in each local extrinsic patch (predefined local unit disks distributed over manifold). Then by taking into the pre-calculated Zernike bases on each local patch, an overdetermined system of linear equations can be solved, which outputs the Zernike coefficients as the extracted intrinsic \lq{patch}\rq.

\paragraph{Zernike convolution} layer takes the extracted intrinsic \lq{patch}\rq outputted from the \textit{Zernike-patch extraction} layer as input. By specifying the number of filters $P$ and the number of all possible rotations $Q$ in a range of $2\pi$ of individual filter, the operation described in (\ref{eq:Zernike convolution 4}) is conducted in this layer with $\triangle\theta = 0, \frac{2\pi}{P},...,\frac{2\pi(P-1)}{P}$, which outputs $P*Q$ dimensional vectors as the input of \textit{Angular max-pooling} layer.

\paragraph{Angular max-pooling} layer is an internal embedded layer in conjunction with the \textit{Zernike convolution} layer, which takes the maximum over $Q$ filter rotations and outputs the evolved geometric feature vectors in a $P$ dimensional space.

\paragraph{Feature-patch normalization } layer is an optional option that can be embedded into the \textit{Zernike-patch extraction} layer, which is used to normalize the partitioned feature vectors distributed in each local patch by subtracting the feature vector located at patch center. Then the normalized vectors in each local patch are used for deriving the Zernike coefficients.

\paragraph{Patch linear/ regression} layer is used to linear transform the geometric feature vector to match a desired output dimensions. For a regression purpose, each geometric feature vector is regressed to a scalar that match the target output at corresponding position. Across the patches, an identical linear transformation is shared.   

For the implementation of Z-CNNs, we use Keras \cite{chollet2015keras} and Tensorflow \cite{tensorflow2015-whitepaper}. The Conv1D layer in Keras is utilized as our \textit{Patch linear/regression} layer due to its weights sharing mechanism. The rest of aforementioned layers are implemented as customized Keras layers to meet our needs.

\subsection{Comparison to other geometric ConvNets}
\label{Comparison}
In previous efforts for defining the convolution on manifolds, an interpolating approach based on spatial neighbors was commonly used for patch extraction. By defining a set of weighting functions $w_1(x,\cdot),...,w_J(x,\cdot)$, the function $f$ at the position $x$ can be interpolated by its neighbors spatially centered at $x$. In this manner, the local patch can be obtained by mapping from the manifold into some local system of coordinates around $x$:
\begin{equation}
    D_j(x)f = \int_{\chi}w_j(x,x')f(x') dx',j=1,...,J.
    \label{eq:patch operator}
\end{equation}
Then a spatial definition of the convolution on manifold is given by following a template-matching procedure:
\begin{equation}
    (f\star g)(x) = \sum_{j=1}^{J} g_j D_j(x)f,
    \label{eq:spatial convolution}
\end{equation}
where $g$ denote the filters (template coefficients) applied on the extracted patch at position $x$. Geodesic CNN \cite{masci2015geodesic}, Anisotropic CNN \cite{boscaini2016learning} and MoNet \cite{monti2017geometric} those pioneers, also considered as state-of-the-art methods in defining convolution on manifold all adopted 
such weights interpolating mechanism, but with different approaches in defining the weighting functions.

Compared to above methods, other than adopting an approach of using surrounding neighbors to interpolating $f$ defined on a local patch by introducing any weighting functions, we introduce a function decomposition approach by using a sequence of orthogonal bases (Zernike functions). In terms of parameterization, our approach usually comes with a less complexity. The number of parameters required for those charting-based methods is equal to the number of the weighting functions, which essentially has to be the number of discretized samples in a local patch. Whereas in our case, the required number of parameters is the number of the orthogonal bases we chose for the decomposition, which practically can be much smaller than the number of discretized samples. 
\section{Case study - wall stress estimation in aneurysm dataset}
\begin{figure}[h]
    \centering
    \includegraphics[width=\textwidth]{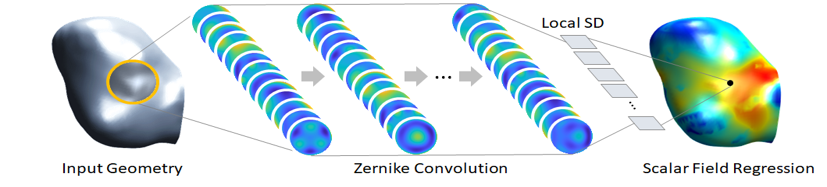}
    \caption{Schematic diagram of ZerNet-based scalar field regression. For each point on the manifold, local Zernike patch is approximated (yellow circle on the left), which then undergoes the Zernike convolution process.}
    \label{Case study schematic diagram}
\end{figure}

 As a case study, we construct a ZerNet to learn the correlations between geometric features and wall stress distribution in cerebral aneurysms. Once the training is completed, by inputting the 3D mesh of any individual cerebral aneurysm, our ZerNet model can estimate the magnitude of wall stress distributed over the mesh surface. Figure \ref{Case study schematic diagram} shows a schematic diagram of our approach for such scalar field regression type of problems.
\subsection{Data set}
Our data set contains $26$ cerebral aneurysm models that modeled as triangular mesh. For each individual model, the wall stress distributed on each of the mesh vertices is provided. All the $26$ mesh models have different mesh topologies. For the sake of obtaining a better reference in setting the radius $r_0$ of the geodesic ball for the local patch extraction (mentioned in section \ref{Zernikeconvolution}) across all models, we normalize all $26$ mesh models to have an identical mesh surface area. Empirically, we choose a sufficiently small $r_0$ to extract each local patch with an area approximately $1\%$ of the total mesh surface area. In order to resolve the uncomformity issue across the mesh models in our data set, we uniformly samples $8000$ random points on each mesh surface by using a sampling algorithm for arbitrary surface in \cite{gptoolbox}. The coordinates of each sampled point is obtained by an interpolation of the three surrounding face vertices on the original mesh. By using the same interpolating weights, the wall stress distributed on the sampled points can be derived. 

\subsection{Input data processing}
For the input data, other than using any sophistic hand-craft geometry vectors, we tend to use more trivial ones to demonstrate our ZerNet's value in discovering latent geometric patterns. The XYZ-coordinates (probability the most standard one in representing any 3D graphic model) of sampled points of each individual aneurysm model are used as the initial input geometric vectors. For initial \textit{Zernike-patch extraction}, after those XYZ-coordinate vectors being partitioned to the positions in each local patch and going through the standard \textit{Feature-patch normalization} process, the normalized vectors within each local patch are further operated by aligning their Z-axis with the normal vector at the patch center to achieve a position and orientation invariant representation of each input aneurysm model. For the Zernike bases defined in each of the local patch, we used the first $21$ normalized Zernike polynomials ($m$ is up to 5 in (\ref{eq:noramlized Zernike func})) calculated by using an implemented algorithm in \cite{fricker2008analyzing}.
\subsection{Results}
\begin{table}[h]
  \centering
  \begin{tabularx}{\textwidth}{l|rrrr|rrrr}
    \toprule
     & \multicolumn{4}{c|}{ZerNet} & \multicolumn{4}{c}{ACNN} \\
    Model ID & MAPE & PCC & HR($10\%$) & HR($20\%$) & MAPE & PCC & HR($10\%$) & HR($20\%$)\\
    \midrule
    TPIa105I & 13.13\% & 0.88 & 58.55\% & 85.25\% & 18.51\% & 0.71 & 41.45\% & 70.33\%\\
    TPIa166I & 10.75\% & 0.85 & 61.47\% & 90.07\% & 21.53\% & 0.64 & 35.19\% & 61.47\%\\
    TPIa182I & 15.43\% & 0.84 & 43.62\% & 75.65\% & 23.37\% & 0.75 & 28.75\% & 53.59\%\\
    TPIa32I  & 17.45\% & 0.85 & 50.02\% & 78.34\% & 32.17\% & 0.68 & 26.53\% & 48.14\%\\
    TPIa33I  & 9.25\% & 0.89  & 68.42\% & 88.86\% & 21.08\% & 0.58 & 35.65\% & 58.38\%\\
    \bottomrule
  \end{tabularx}
  \caption{Performance of ZerNet vs ACNN over the five validation aneurysm models. Performance was quantitative evaluated with following criteria: mean absolute percentage error (MAPE), Pearson correlation coefficient (PCC) and Hit-rate (HR). By restricting absolute percentage error within a tolerance threshold ($10\%$ and $20\%$), Hit-rate was calculated as the percentage ratio of the number of the vertices that have the wall stress precisely estimated over the total number of mesh vertices of each aneurysm model.}
  \label{Prediction-table}
\end{table}
We constructed a ZerNet architecture containing three \textit{Zernike convolution} layers with the number of filters setting as $128$, $512$ and $1024$ respectively. For a further evolving of the geometric feature vectors, we appended one more \textit{Patch linear} layer with a setting of $800$ as output dimension in conjunction with the last \textit{Zernike convolution} layer. Finally a \textit{Patch regression} layer with output dimension $1$ was set as the output layer. Among the $26$ aneurysm models, we randomly selected $5$ models for performing leave-one-out cross validations. For each of the model among the $5$, we left it out as the test/validation set and used the rest $25$ models in our data set to train the ZerNet. Training was performed by using Adam optimizer \cite{DBLP:journals/corr/KingmaB14} to minimize mean squared error between the network output and the provided wall stress distribution of each trained model.

\begin{figure}[ht]
    \centering
    \includegraphics[width=\textwidth]{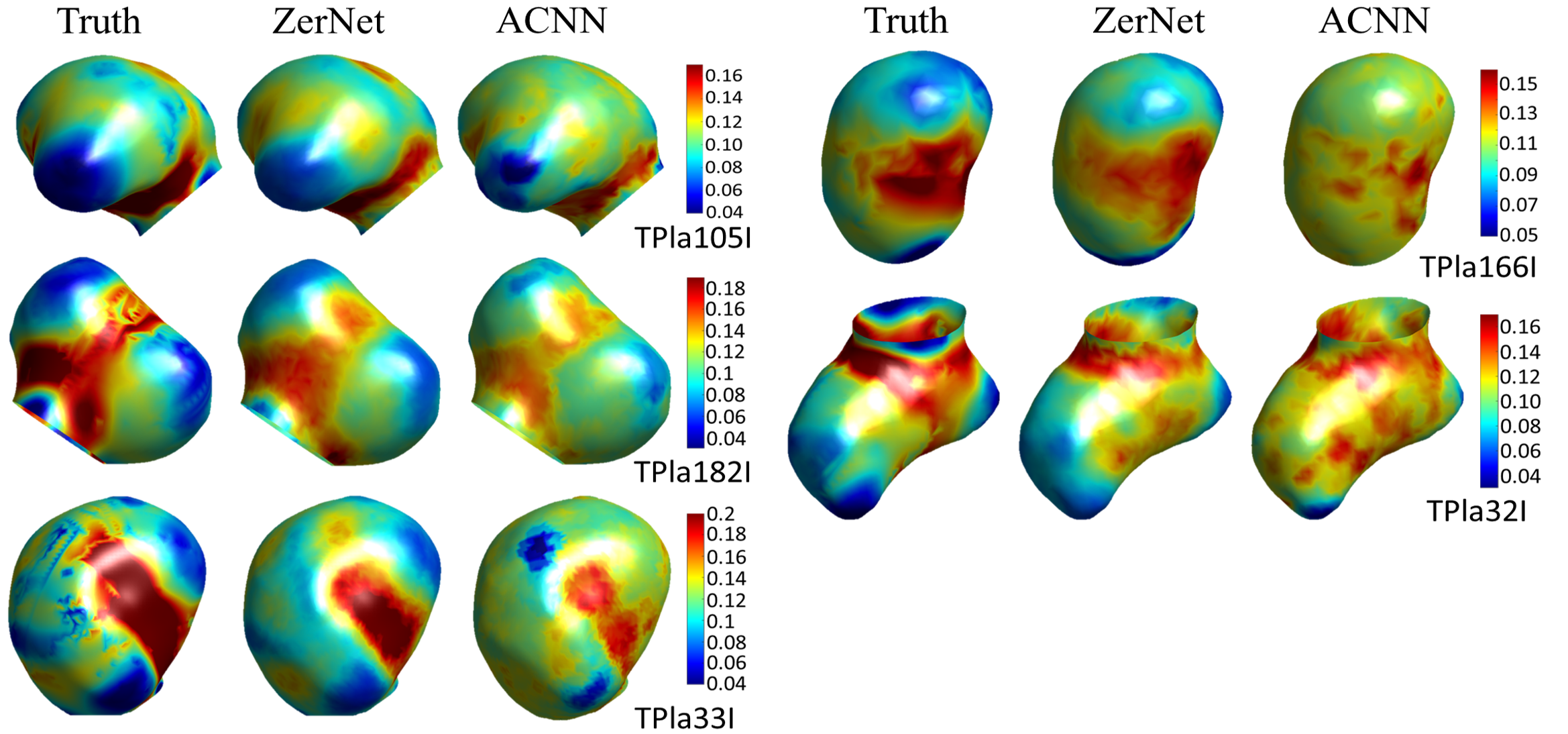}
    \caption{Visualization of estimated wall stress distribution range maps on the five validation aneurysm models. For each of the aneurysm models, the true wall stress distribution is shown on the left as reference.}
    \label{fig: vis_result}
\end{figure}

For the performance evaluation, we compared our method to current state-of-the-art method, Anisotropic Convolutional Neural Networks (ACNN) \cite{boscaini2016anisotropic}. Due to the requiring of a consistent triangular mesh topology across the models as for ACNN method, we used the original mesh of one aneurysm model as template to register the rest $25$ aneurysm models by using a deformable registration algorithm \cite{li08global}. For the input of the ACNN, we used the local SHOT descriptor \cite{salti2014shot} with $544$ dimensions as suggested in \cite{bronstein2017geometric}. For the architecture of ACNN, we constructed three Anisotropic convolutional layers with setting the same number of filters for each as in ZerNet; other layers are equivalently set as in ZerNet as well. Finally, for each validation model, the predicted wall stress outputted by ZerNet and ACNN were both mapped back to every vertex on the original triangular mesh via nearest neighbors search. The quantitative evaluation of our method compared to ACNN was summarized in Table \ref{Prediction-table}. Figure \ref{fig: vis_result} shows qualitative visualizations of the wall stress distribution estimated by our method and ACNN. We can clearly see that our proposed ZerNet has a promising performance in such aneurysm wall stress estimation task, which significantly outperforms ACNN.
\section{Conclusion}
In this paper, we introduced a new concept of Zernike convolution by generalizing the convolution operation to surface manifolds. Building upon such, we proposed the novel ZerNet architecture, which extends conventional ConvNets to learn geometric features from 3d graphic models. Experiments has shown that the ZerNet were indeed capable of capturing the local geometry features correlated to the defined task (wall stress on cerebral aneurysms), showing a promising estimation of the stress distribution trend. For the definition of such ZerNet, no problem-specific assumptions has been made, such that many other problems of a similar kind can benefit from our new method. 

To our knowledge, our work is the first to generalize ConvNets in attempts for local-features regression within manifolds domain. Unlike the many classification based problems that have been addressed by previous works, which are challenging for computer but relatively perceptually easy for humans such as finding shape correspondence, shape retrieval and \textit{etc}, our work acts as a paradigm in a sense that how the "geometric deep learning" can really facilitate the solving for those perceptually difficult tasks.

\end{document}